\documentclass[runningheads]{llncs}
\usepackage[T1]{fontenc}
\usepackage{amsmath,amssymb,amsfonts}
\usepackage{graphicx}
\usepackage{multirow}
\usepackage{booktabs}
\usepackage{subfigure}
\begin{document}
\title{OEL-AD: An Online Evolutive Learning Framework for Cross-Region Adaptive Autonomous Driving}
%
%
\author{Jiayue Jin\inst{1,2} \and
Lang Qian\inst{1} \and
Jingyu Zhang\inst{1} \and
Chuanyu Ju\inst{1} \and
Liang Song\inst{1,2,*}
}
\authorrunning{J. Jin et al.}
\titlerunning{OEL-AD}
%
\institute{College of Intelligent Robotics and Advanced Manufacturing, Fudan University, Shanghai, China \and
Innovation Platform for Academicians of Hainan Province, Haikou, China
\email{24110860008@m.fudan.edu.cn, songl@fudan.edu.cn}}
\maketitle              
\begin{abstract}
Autonomous driving models have achieved great progress in recent years, yet most existing approaches remain static, following a train-then-deploy paradigm that lacks the ability to adjust internal parameters when encountering new environments. This limitation is particularly critical in cross-region scenarios, which are common in real-world deployments in autonomous driving where differences in road layouts, traffic patterns, and environmental conditions introduce distribution shifts. In this paper, we argue for a paradigm shift towards Online Evolutive Learning (OEL), a learning paradigm designed to enable adaptive, online model optimization through the interaction and coordination of intelligent components in dynamic environments. We introduce OEL-AD, a novel framework that instantiates this paradigm for autonomous driving. Specifically, OEL-AD leverages trajectory uncertainty from the planning module to determine whether adaptive updates are necessary. When triggered, the framework uses discrepancies between predicted agent behaviors and subsequent observations as a self-supervised learning signal to refine perception and prediction modules, thereby indirectly improving planning accuracy and safety. Experiments on the nuScenes dataset with a cross-region split between Singapore and Boston demonstrate that OEL-AD consistently improves perception and prediction performance. This enhancement in perception and prediction consequently results in improved planning capabilities, thereby confirming the promise of online evolutive learning as a fundamental approach for building adaptive and robust autonomous driving systems.

\end{abstract}
\section{Introduction}
Autonomous driving has become one of the most influential research areas in artificial intelligence and robotics, supported by advances in perception, prediction, and planning algorithms, large-scale datasets, and increasingly powerful computing platforms. Recent progress has demonstrated a clear trend from modular pipelines toward more integrated end-to-end architectures \cite{chen2024end}. End-to-end models are able to jointly optimize multiple tasks, benefiting from shared representations and improved coordination between modules, and have achieved strong results on curated datasets and established benchmarks. 

Despite these achievements, most current methods follow a static train-then-deploy paradigm, in which models are optimized offline on fixed datasets and deployed as frozen systems. This paradigm assumes that training and deployment environments are drawn from similar distributions, an assumption that often does not hold in real-world conditions. Autonomous driving systems must operate in environments that are highly diverse and dynamic, influenced by regional differences, weather changes, and evolving traffic conditions. Under such circumstances, models trained in one environment frequently experience performance degradation when deployed in new contexts, highlighting the limitations of static pipelines.

Among them, cross-region generalization represents one of the most challenging and common scenarios in real-world autonomous driving. Variations between cities in terms of road topology, traffic density, and driving behavior create distribution shifts that are difficult to capture during offline training. Addressing such distribution shifts is crucial for building reliable and scalable autonomous driving systems. Existing approaches to domain adaptation \cite{shan2019pixel,li2022cross,li2023domain,yasarla2025roca} attempt to alleviate these issues through additional labeled data, domain-invariant representations, or adversarial learning strategies. While effective to some extent, these approaches are fundamentally constrained by their dependence on pre-defined training datasets and lack the ability to dynamically adapt during deployment \cite{farahani2021brief}. Recent research on test-time adaptation \cite{sun2020test,yuan2023robust,segu2023darth} demonstrates the promise of updating models on-the-fly using unlabeled data; however, these methods typically focus on single-task perception, operate in isolation from downstream planning, limiting their long-term effectiveness for safety-critical applications.

To address these limitations, we argue for a paradigm shift to Online Evolutive Learning (OEL) \cite{song2022networking}, a learning paradigm that enables autonomous driving systems to refine internal representations and policies in real time through a closed-loop optimization process. Instead of treating perception, prediction, and planning as isolated stages, OEL promotes collaboration between modules, allowing the system to selectively update its components based on feedback from decision-making outcomes. By incorporating adaptation into the system’s operational cycle, OEL reduces the dependence on curated training datasets and improves the ability to handle dynamic, open-world environments.

Building on this paradigm, we instantiate OEL in the autonomous driving domain by introducing the OEL-AD framework. OEL-AD uses an uncertainty-triggered adaptation mechanism, where the confidence of the system’s planning outputs is used to determine when updates are necessary. An attention-guided module focuses adaptation efforts on the most influential agents in the driving scene, and a self-supervised optimization strategy based on temporal consistency adjusts perception and prediction models by comparing predicted trajectories with subsequent observations. This feedback-driven design enables the autonomous driving system to perform targeted adaptation without requiring additional labeled data, thus improving robustness.

We evaluate OEL-AD on the nuScenes dataset \cite{caesar2020nuscenes} under a challenging cross-region setting between Singapore and Boston. The experiments demonstrate that OEL-AD consistently improves perception and prediction performance, leading to better planning accuracy and enhanced safety under distribution shifts. The main contributions are as follows:
\begin{itemize}
    \item We propose OEL-AD, a novel framework that instantiates the OEL paradigm in autonomous driving. Unlike conventional train-then-deploy approaches, OEL-AD enables models to perform online adaptation during deployment, effectively handling cross-region distribution shifts.
    
    \item We design an uncertainty-aware and attention-guided update mechanism that leverages trajectory uncertainty to selectively trigger model evolution. By focusing only on high-uncertainty and high-attention objects, OEL-AD performs efficient self-supervised updates to the perception and prediction modules without requiring additional human annotations.
    
    \item We conduct extensive cross-region evaluations on the nuScenes dataset, simulating realistic deployment scenarios where models are trained in one city and tested in another. Results demonstrate that OEL-AD consistently improves both perception and prediction accuracy under challenging domain shifts, leading to enhanced planning reliability and lower collision rates. These findings validate OEL-AD as a promising step towards adaptive autonomous driving systems.
\end{itemize}

\section{Related Works}
\subsection{Autonomous Driving Models}

Autonomous driving systems have evolved significantly, progressing from traditional modular architectures to modern end-to-end models. Traditional systems, which sequentially connect perception, prediction, and planning, offer strong interpretability and are easier to diagnose. Nevertheless, this pipeline design is prone to accumulating errors between sequential modules \cite{chen2024end}. To mitigate this limitation, recent studies increasingly employ end-to-end solutions \cite{hu2022stp3,hu2023uniad,jiang2023vad,chen2024vadv2,li2024hydra,sun2024sparsedrive}. Such differentiability facilitates joint training across tasks, which enhances overall system performance. Our proposed framework is designed to be broadly compatible, supporting both traditional modular and modern end-to-end architectures. Furthermore, planning modules within these models have also advanced. Most contemporary planners generate multiple trajectories with corresponding probabilities to capture environmental uncertainty \cite{chen2024vadv2,li2024hydra,sun2024sparsedrive}, offering greater robustness than earlier approaches that produced only a single deterministic trajectory \cite{hu2022stp3,hu2023uniad,jiang2023vad}. Our work builds upon this capability. Concurrently, transformer-based designs have become prevalent in planning modules, utilizing attention mechanisms to enhance decision-making \cite{chitta2022transfuser,jia2025drivetransformer}. We capitalize on this common architectural feature by using the attention weights to identify the most influential objects for our self-supervised updates. Overall, OEL-AD is highly generalizable and supports the majority of contemporary autonomous driving models.

\subsection{Cross-Region Adaptive Autonomous Driving Methods}

While the majority of research in autonomous driving follows the paradigm of proposing novel architectures, training them on a designated training set, and validating on a test set, a growing body of work has begun to address the generalization and adaptation of models when facing distribution shifts especially the cross-region scenarios. One category of research leverages foundation models to create autonomous driving architectures \cite{pan2024vlp}. These methods capitalize on the zero-shot generalization capabilities of large models, which can yield strong performance across a variety of environments. However, the large models themselves remain static; they may still produce decision-making errors when confronted with drastically changing environments, and the cost of adapting to each new domain is substantial. Domain adaptation (DA) techniques are also explored to address regional distribution shifts \cite{shan2019pixel,li2022cross,li2023domain,yasarla2025roca}. However, these approaches typically introduce additional training stages and rely on extra labeled data, which limits their efficiency in real-world deployment. A related category includes test-time adaptation (TTA) strategies \cite{sun2020test,yuan2023robust,segu2023darth,sima2025centaur,yang2025codemerge}. Unlike the aforementioned approaches, the Online Evolutive Learning (OEL) method we propose is a lightweight framework that enables the model to evolve and adapt in real-time during deployment, offering a more efficient solution to handling distribution shifts.

\subsection{Online Evolutive Learning}
Online Evolutive Learning (OEL) is a learning paradigm designed to enable adaptive, online model optimization through the interaction and coordination of multiple intelligent components, such as perception, control, and generation, within dynamic environments \cite{song2022networking}. OEL emphasizes real-time learning and adaptation based on agent-environment interaction and inter-agent collaboration. Several recent works have explored this direction for general multi-agent systems. For example, the Collaborative Normality Learning framework \cite{liu2022collaborative} proposes a structure for continual adaptation in multi-agent systems. Mutual Match (MM) \cite{li2023mutual} explores information consistency between agents for robust distributed learning. Additionally, Qian et al. \cite{qian2024new} propose a bidirectional optimization framework that integrates supervised learning and reinforcement learning agents, enabling adaptation through coordinated sensing and control. While these works focus on OEL for general agents, our paper applies this paradigm specifically to the domain of autonomous driving. We enable the model itself to evolve online, using feedback to trigger adaptation and leveraging attention mechanisms to guide self-supervised updates to improve robustness against cross-region distribution shifts.

\begin{figure}
\begin{center}
\includegraphics[width=0.8\textwidth]{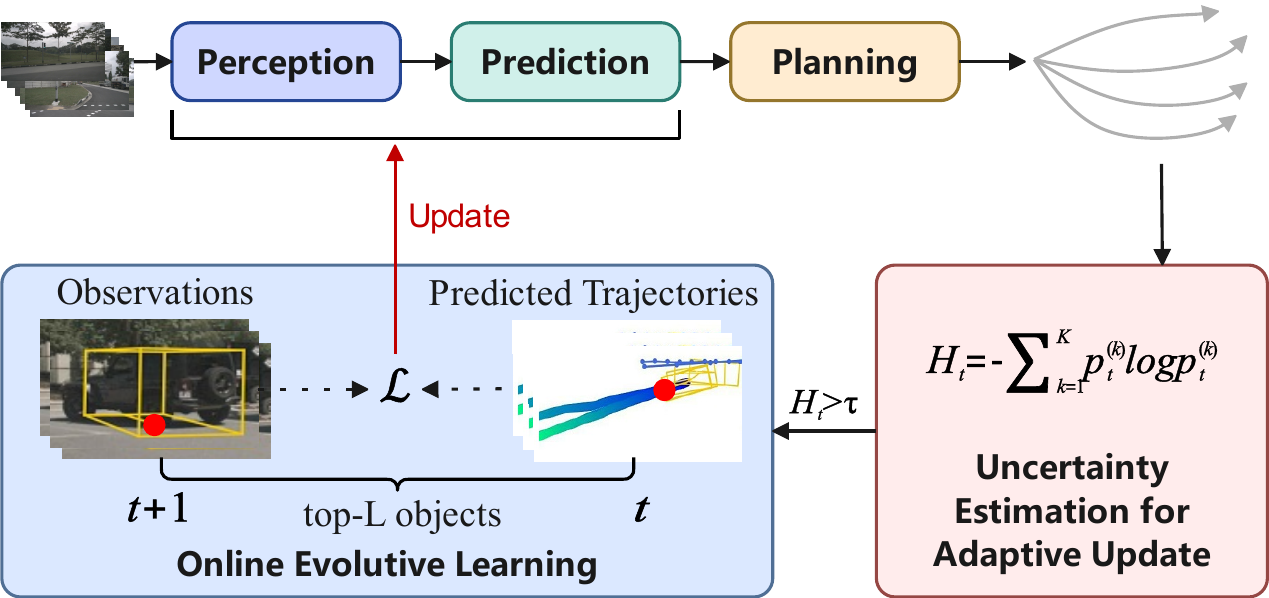}
\end{center}
   \caption{Overview of OEL-AD. The framework maps raw observations to multi-trajectories, measures trajectory uncertainty, and triggers an online, self-supervised update when $H_t>\tau$. The selected objects provide next timestep supervisions to refine perception and prediction, thereby enhancing planning. }
\label{fig:overview}
\end{figure}

\section{Method}
We introduce our proposed OEL-AD framework in this section, which is built upon the Online Evolutive Learning (OEL) paradigm to enable adaptation in cross-region scenarios. The OEL-AD framework is presented in Fig.~\ref{fig:overview}, providing a high-level overview of its core components and their interactions. OEL-AD integrates perception, prediction, and planning into a closed-loop system where trajectory uncertainty is used to determine the need for online updates, and self-supervised signals from future observations guide targeted adaptation. 
\subsection{Forward Inference}

The forward inference phase is a necessary step during model deployment. Since our OEL-AD is based on the results of forward inference, we first describe this process for completeness, covering perception, prediction, and trajectory planning. During inference, the trained driving model directly transforms sensor observations into multiple candidate motion plans for the ego vehicle. Let $o_t$ denote the raw sensor inputs at time step $t$. An encoder $f_\theta$ maps these observations into a latent representation of the scene:
\begin{equation}
z_t = f_\theta(o_t), \quad o_t \in \mathcal{O}, \; z_t \in \mathbb{R}^d,
\end{equation}
where $z_t$ encodes road context, dynamic agent states, and their potential future evolutions.

Based on this latent representation, the model implicitly integrates prediction of surrounding agents' behaviors through an internal transformation $h_\psi(z_t)$. The resulting predictive features are fused back into the latent representation, yielding
\begin{equation}
\tilde{z}_t = \Phi(z_t, h_\psi(z_t)),
\end{equation}
where $\Phi(\cdot)$ denotes the fusion operation that enriches the scene representation with predictive information.

The planning head $g_\phi$ then generates a set of $K$ candidate trajectories for the ego vehicle over a horizon of length $H$, together with a probability for each candidate:
\begin{equation}
\big(\hat{X}_t, \pi_t\big) = g_\phi(\tilde{z}_t),
\end{equation}
\begin{equation}
\hat{X}_t = \left\{\hat{x}_t^{(k)}\right\}_{k=1}^K, \quad 
\hat{x}_t^{(k)} = \left\{\hat{\text{pt}}_{t+1}^{(k)}, \hat{\text{pt}}_{t+2}^{(k)}, \dots, \hat{\text{pt}}_{t+H}^{(k)}\right\},
\end{equation}
\begin{equation}
\pi_t = \left\{\pi_t^{(k)}\right\}_{k=1}^K, \quad \pi_t^{(k)} \in [0,1], \quad \sum_{k=1}^{K} \pi_t^{(k)} = 1,
\end{equation}

Thus, the overall inference process can be summarized as
\begin{equation}
\big(\hat{X}_t, \pi_t\big) = g_\phi\Big(\Phi\big(f_\theta(o_t), h_\psi(f_\theta(o_t))\big)\Big),
\end{equation}
which describes the mapping from raw observations to a multi-trajectory output with associated probabilities.

\subsection{Uncertainty Estimation for Adaptive Update}

To determine whether an adaptive update is required, the system evaluates the uncertainty of the candidate planning trajectories. To be specifically, the planning module produces $K$ trajectories with associated probability scores $\pi_t = \{\pi_t^{(1)}, \pi_t^{(2)}, \dots, \pi_t^{(K)}\}$. These scores are first normalized through a softmax operation to obtain a valid probability distribution:
\begin{equation}
p_t^{(k)} = \frac{\exp(\pi_t^{(k)})}{\sum_{j=1}^{K} \exp(\pi_t^{(j)})}, \quad \sum_{k=1}^{K} p_t^{(k)} = 1.
\end{equation}
The entropy of this distribution is then computed as
\begin{equation}
H_t = - \sum_{k=1}^{K} p_t^{(k)} \log p_t^{(k)},
\end{equation}
which measures the degree of uncertainty in the model’s trajectory selection.  

A predefined threshold $\tau$ is used to decide whether the uncertainty level indicates potential distribution shift. If the entropy exceeds the threshold, the system triggers an adaptive update procedure:
\begin{equation}
H_t > \tau \;\;\; \Rightarrow \;\;\; \text{perform online evolutive learning},
\end{equation}
otherwise the model continues inference without update:
\begin{equation}
H_t \leq \tau \;\;\; \Rightarrow \;\;\; \text{skip adaptive update}.
\end{equation}
The threshold $\tau$ can be calibrated on a validation set by measuring the average entropy of planning outputs under in-distribution scenarios. 

\subsection{Online Evolutive Learning}

When the entropy $H_t$ of the trajectory distribution exceeds the threshold $\tau$, the system activates the online evolutive learning procedure. The goal is to refine perception and prediction capabilities in order to enhance the planning performance under cross-region distribution shifts.

First, from the planning module, the ego vehicle’s attention scores over surrounding objects are obtained. Let $\mathcal{A}_t$ denote the set of indices corresponding to the L highest-ranked agents, determined by their attention scores. For each selected object $i \in \mathcal{A}_t$, the prediction module provides a set of future trajectories $\hat{y}_t^{(i)} = \{\hat{\text{pt}}_{t+1}^{(i)}, \hat{\text{pt}}_{t+2}^{(i)}, \dots, \hat{\text{pt}}_{t+H}^{(i)}\}$. The trajectory point of interest is the immediate next-step prediction $\hat{\text{pt}}_{t+1}^{(i)}$. At the next time step, the perception module provides an updated observation of object $i$, typically represented as a bounding box that includes position, size, orientation, and velocity. From this representation, the ground-truth center position of the object is extracted and denoted as $\text{pt}_{t+1}^{(i)}$.

The prediction error is then measured as the discrepancy between the predicted and observed positions:
\begin{equation}
\mathcal{L}_t^{(i)} = \big\| \hat{\text{pt}}_{t+1}^{(i)} - \text{pt}_{t+1}^{(i)} \big\|_2^2, \quad i \in \mathcal{A}_t.
\end{equation}
The overall online evolutive learning loss aggregates these errors across the attended objects:
\begin{equation}
\mathcal{L}_t = \frac{1}{|\mathcal{A}_t|} \sum_{i \in \mathcal{A}_t} \mathcal{L}_t^{(i)}.
\end{equation}
Finally, this loss is used to update the parameters of both the perception module $f_\theta$ and the prediction module $h_\psi$:
\begin{equation}
\theta \leftarrow \theta - \eta \nabla_\theta \mathcal{L}_t, \quad 
\psi \leftarrow \psi - \eta \nabla_\psi \mathcal{L}_t,
\end{equation}
where $\eta$ is the learning rate. By continually refining these modules online, the system improves its ability to perceive dynamic agents and predict their future motions, which in turn enhances the reliability of the planning outputs. This procedure naturally forms a self-supervised online update loop, since the perception outputs at the next timestep serve as immediate supervision for the prediction errors. As a result, the model can adapt to unseen cross-region environments without requiring additional manual annotations.

\section{Experiments}
\subsection{Experimental Setup}
\subsubsection{Datasets.}
We use the nuScenes dataset \cite{caesar2020nuscenes} to conduct experiments. It contains diverse traffic scenarios, including multiple agents such as vehicles, pedestrians, and cyclists, as well as rich sensor information from LiDAR, cameras, and radar. To evaluate the model's ability to handle cross-region distribution shifts, we split the dataset into two geographic regions, Singapore and Boston. One region is used for training and the other for testing, and vice versa.

\subsubsection{Implementation Details.}
We adopt SparseDrive \cite{sun2024sparsedrive} as our base model and extend it with the proposed OEL-AD framework. During the training phase, we use four 24GB NVIDIA RTX 4090 GPUs. For the OEL stage, which is performed during inference when the uncertainty exceeds the threshold, we use one 24GB NVIDIA RTX 3090 GPU. We use a learning rate of $3 \times 10^{-7}$ for online updates and select the top 35 attended objects ($L=35$) for loss computation.


\subsubsection{Evaluation Metrics.}
We evaluate the proposed framework across perception, prediction, and planning components using widely adopted metrics. For perception, we use nuScenes Detection Score (NDS) to evaluate the model performance. For prediction, following the standard motion prediction protocols, we adopt conventional metrics, including minADE (minimum Average Displacement Error), minFDE (minimum Final Displacement Error) and EPA (End-to-end Prediction Accuracy). For planning, we adopt commonly used L2 error and collision rate to evaluate the planning performance, aligning with SparseDrive \cite{sun2024sparsedrive}. Together, these metrics provide a comprehensive assessment of the system’s ability to accurately perceive its surroundings, forecast future dynamics, and produce safe and reliable trajectories under cross-region distribution shifts.

\subsection{Experimental Results}
\begin{table}
\centering
\renewcommand{\arraystretch}{1.2}
\setlength{\tabcolsep}{3pt}
\caption{Cross-region planning results of four methods (UniAD \cite{hu2023uniad}, VAD \cite{jiang2023vad}, SparseDrive \cite{sun2024sparsedrive}, OEL-AD). 
Metrics are reported for prediction horizons of 1s, 2s, and 3s. Lower L2 and Collision Rate values indicate better performance. OEL-AD consistently achieves the best overall results across both scenarios.
Abbreviations: SG = Singapore, BOS = Boston.}
\label{tab:cross-region-planning}

\begin{tabular}{c|c|ccc|ccc}
\toprule
\multirow{2}{*}{\textbf{Scenario}} & \multirow{2}{*}{\textbf{Methods}} 
& \multicolumn{3}{c|}{\textbf{L2 (m)$\downarrow$}} 
& \multicolumn{3}{c}{\textbf{Collision Rate (\%)$\downarrow$}} \\
& & \textbf{1s} & \textbf{2s} & \textbf{3s} 
& \textbf{1s} & \textbf{2s} & \textbf{3s} \\
\midrule
\multirow{4}{*}{\textbf{SG$\rightarrow$BOS}} 
& UniAD        & 0.4069 & 0.8403 & 1.9228 & 0.038 & 0.121 & 0.958 \\
& VAD          & 0.4175 & 0.8621 & 2.2082 & 0.035 & 0.119 & 0.516 \\
& SparseDrive  & 0.4057 & 0.8350 & 1.4294 & 0.034 & 0.117 & 0.459 \\
& OEL-AD      & \textbf{0.3904} & \textbf{0.8054} & \textbf{1.3645} & \textbf{0.032} & \textbf{0.117} & \textbf{0.377} \\
\midrule
\multirow{4}{*}{\textbf{BOS$\rightarrow$SG}} 
& UniAD        & 0.6017 & 1.1601 & 1.9652 & 0.055 & 0.256 & 0.836 \\
& VAD          & 0.6019 & 1.1611 & 1.9970 & 0.061 & 0.258 & 0.978 \\
& SparseDrive  & 0.6001 & 1.1567 & 1.8534 & 0.065 & 0.283 & 0.797 \\
& OEL-AD      & \textbf{0.5909} & \textbf{1.1429} & \textbf{1.8361} & \textbf{0.054} & \textbf{0.256} & \textbf{0.753} \\
\bottomrule
\end{tabular}
\end{table}



\begin{table}
\centering
\renewcommand{\arraystretch}{1.2}
\setlength{\tabcolsep}{4pt} 
\caption{Cross-region perception (NDS) and prediction (minADE, minFDE, EPA) results of SparseDrive \cite{sun2024sparsedrive} and OEL-AD across SG$\rightarrow$BOS and BOS$\rightarrow$SG. Higher is better for NDS and EPA, while lower is better for minADE and minFDE. OEL-AD consistently achieves the best overall results across both scenarios. Abbreviations: SG = Singapore, BOS = Boston.}

\label{tab:cross-region-perception}

\begin{tabular}{c|c|c|ccc}
\toprule
\multirow{2}{*}{\textbf{Scenario}} & 
\multirow{2}{*}{\textbf{Methods}} & 
\textbf{Perception} & 
\multicolumn{3}{c}{\textbf{Prediction}} \\
& & \textbf{NDS$\uparrow$} & \textbf{minADE$\downarrow$} & \textbf{minFDE$\downarrow$} & \textbf{EPA$\uparrow$} \\
\midrule
\multirow{2}{*}{\textbf{SG$\rightarrow$BOS}} 
& SparseDrive  & 0.2453 & 0.9669 & 1.4718 & 0.2820 \\
& OEL-AD       & \textbf{0.2454} & \textbf{0.9565} & \textbf{1.4596} & \textbf{0.2834}  \\
\midrule
\multirow{2}{*}{\textbf{BOS$\rightarrow$SG}} 
& SparseDrive  & 0.2982 & 1.3696 & 2.3571 & 0.2641 \\
& OEL-AD       & \textbf{0.2993} & \textbf{1.3606} & \textbf{2.3561} & \textbf{0.2660}  \\
\bottomrule
\end{tabular}
\end{table}

\begin{figure*}[t]  
    \centering
    \subfigure[Baseline]{
        \includegraphics[width=0.98\textwidth]{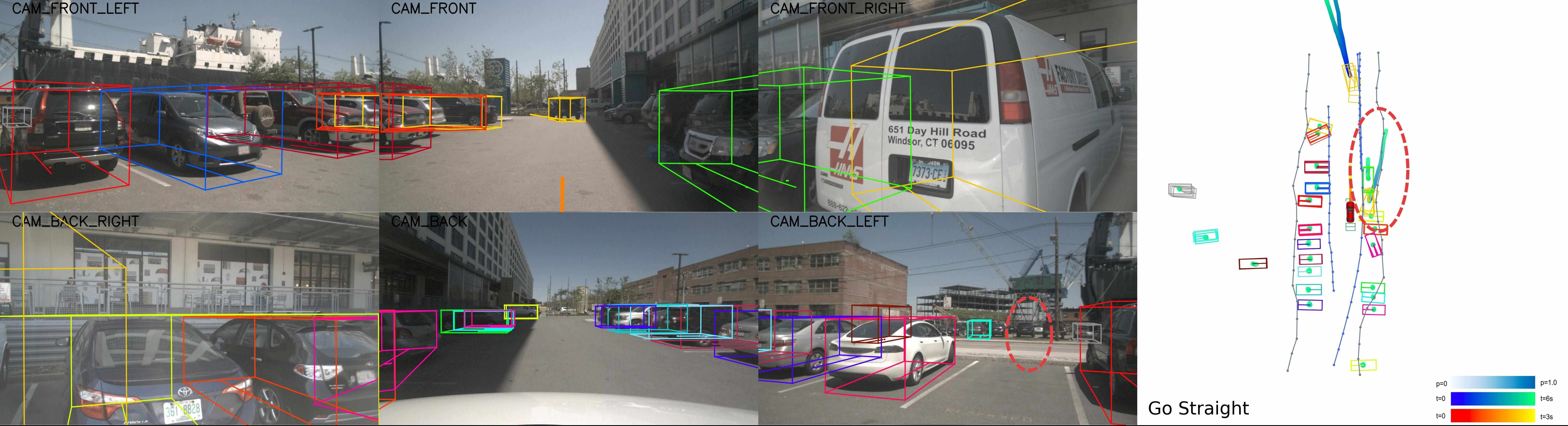}
    }
    \subfigure[OEL-AD]{
        \includegraphics[width=0.98\textwidth]{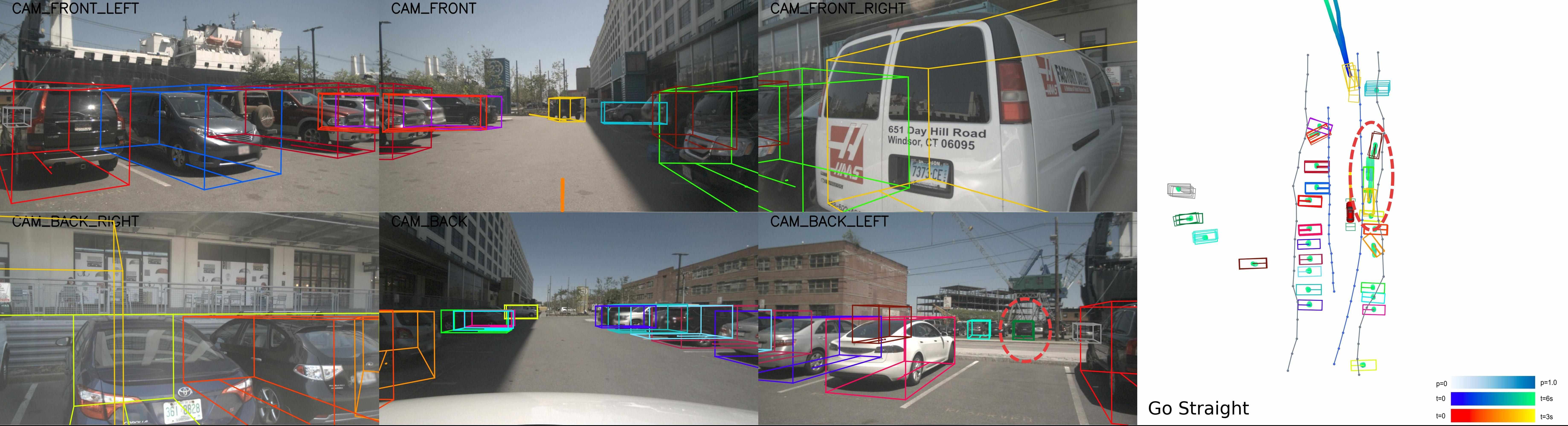}
    }
    \caption{Qualitative results. Compared to the baseline, OEL-AD achieves higher accuracy in detecting distant objects. Moreover, it demonstrates superior prediction accuracy for surrounding agents. In our experiments, the surrounding vehicles are stationary; OEL-AD correctly predicts minimal movement for these vehicles, whereas the baseline erroneously predicts significant motion trajectories for the same agents, which is clearly incorrect.}
    \label{fig:qualitative}
\end{figure*}

\subsubsection{Results on Cross-Region Nuscenes Dataset.}
Tables~\ref{tab:cross-region-planning} and~\ref{tab:cross-region-perception} summarize the quantitative results of different methods on the cross-region nuScenes benchmark, where models are trained on one region (Singapore or Boston) and evaluated on the other to simulate distribution shifts. Across both transfer directions (SG$\rightarrow$BOS and BOS$\rightarrow$SG), OEL-AD consistently outperforms prior approaches, including UniAD, VAD, and SparseDrive, in terms of perception, prediction, and planning metrics. For perception, OEL-AD achieves higher NDS scores than SparseDrive, indicating improved 3D object detection performance under cross-domain distribution shifts. In prediction, OEL-AD reduces both minADE and minFDE across all scenarios, while achieving higher EPA, demonstrating its ability to generate more accurate and reliable trajectory forecasts. In planning, OEL-AD achieves the lowest L2 trajectory error and collision rate at all prediction horizons (1s, 2s, and 3s). For instance, under the SG$\rightarrow$BOS setting, OEL-AD reduces the 3s L2 error to 1.36\,m compared to 1.43\,m for SparseDrive, and lowers the 3s collision rate to 0.377\%, the best among all methods. Similar trends are observed in the BOS$\rightarrow$SG scenario, confirming that the benefits of OEL generalize across domains. These results validate that OEL-AD effectively leverages online self-supervised signals to adapt perception and prediction components in real time, leading to a closed-loop improvement in planning safety and accuracy. By dynamically adjusting to environment-specific characteristics such as traffic density, agent behavior, and scene geometry, OEL-AD demonstrates strong robustness and adaptability for cross-region autonomous driving deployment.

\subsubsection{Qualitative Results}

Fig.~\ref{fig:qualitative} illustrates qualitative examples that demonstrate the adaptive capability of the proposed OEL-AD framework. These results are obtained by training the model on the Singapore subset and performing online evolutive learning on the Boston validation set. Compared to the baseline, OEL-AD more accurately detects distant objects, reflecting improved perception under cross-region distribution shifts. In addition, OEL-AD provides more precise predictions of surrounding agents’ behaviors. For instance, in scenarios where nearby vehicles are stationary, OEL-AD correctly predicts minimal movement, whereas the baseline incorrectly forecasts significant motion trajectories for the same agents. This behavior highlights the ability of OEL-AD to leverage online self-supervised signals to adjust predictions based on real-time observations, leading to more reliable and physically consistent outputs.


\section{Conclusion}
We introduced OEL-AD, an adaptive autonomous driving framework built on the OEL paradigm to achieve real-time adaptation under cross-region distribution shifts. Unlike conventional train-then-deploy approaches, OEL-AD continuously monitors trajectory-level uncertainty to determine when online updates are required and leverages discrepancies between predicted agent behaviors and subsequent observations as self-supervised signals to refine perception and prediction modules. This feedback-driven process enhances planning performance and safety by enabling the system to dynamically adjust its internal models without additional labeled data. Extensive experiments on the nuScenes dataset with a challenging cross-region split between Singapore and Boston show that OEL-AD consistently improves perception and prediction accuracy, leading to more reliable planning. These results demonstrate the effectiveness of integrating online optimization and self-supervised learning into a closed-loop system and highlight the potential of OEL as a foundation for building autonomous driving systems capable of operating robustly in diverse and evolving environments. Future research will focus on several directions. First, although OEL-AD successfully adapts to unseen environments, it does not explicitly address catastrophic forgetting. Incorporating continual learning techniques will be essential to ensure the retention of previously acquired knowledge. Second, this study focuses primarily on cross-region distribution shifts; we will extend the evaluation to a broader range of scenarios, including diverse weather conditions and large-scale datasets, to further validate the scalability and generality of the framework. 


\begin{credits}
\subsubsection{\ackname} This work was supported by National Key Research and Development Program of China, Project No.2024YFE0200700,Subject No.2024YFE0200703. This work was also supported in part by the Specific Research Fund of the Innovation Platform for Academicians of Hainan Province under Grant YSPTZX202314, in part by the Shanghai Key Research Laboratory of NSAI and the Joint Laboratory on Networked AI Edge Computing, Fudan University-Changan.
\end{credits}
%
%
%
%

\end{document}